\documentclass{article}
\usepackage{ijcai24}

\usepackage{times}
\usepackage{soul}
\usepackage{url}
\usepackage[hidelinks]{hyperref}
\usepackage[utf8]{inputenc}
\usepackage[small]{caption}
\usepackage{graphicx}
\usepackage{amsmath}
\usepackage{amsthm}
\usepackage{booktabs}
\usepackage{algorithm}
\usepackage{algorithmic}
\usepackage[switch]{lineno}
\usepackage{nolbreaks}
\usepackage{listings}

\newcommand\blfootnote{%
  \begingroup
  \renewcommand\thefootnote{}\footnote{Workshop on Artificial Intelligence for Critical Infrastructure (AI4CI 2024) @ IJCAI'24 , Jeju Island, South Korea, \url{https://sites.google.com/view/aiforci-ijcai24/}, August 4, 2024. Eds: F. Silva, W. Su, R. Glatt, Y. Wang.}%
  \addtocounter{footnote}{-1}%
  \endgroup
}

\title{A Toolbox for Supporting Research on AI in Water Distribution Networks}

\author{
 \nolbreaks {André Artelt$^{1,2}$}
\and
 \nolbreaks {Marios S. Kyriakou$^2$}\and
  \nolbreaks{Stelios G. Vrachimis$^{2,3}$}\and
 \nolbreaks {Demetrios G. Eliades$^{2}$}\and\\
 \nolbreaks{Barbara Hammer$^1$}\And
  \nolbreaks{Marios M. Polycarpou$^{2,3}$}\\
\affiliations
$^1$Faculty of Technology, Bielefeld University, Germany\\
$^2$KIOS Research and Innovation Center of Excellence, University of Cyprus, Cyprus\\
$^3$Department of Electrical and Computer Engineering, University of Cyprus, Cyprus\\
\emails
\{artelt.andre, kiriakou.marios, vrachimis.stelios, eldemet, mpolycar\}@ucy.ac.cy, bhammer@techfak.uni-bielefeld.de
}

\begin{document}

\maketitle

\begin{abstract}
Drinking water is a vital resource for humanity, and thus, Water Distribution Networks (WDNs) are considered critical infrastructures in modern societies.
The operation of WDNs is subject to diverse challenges such as water leakages and contamination, cyber/physical attacks, high energy consumption during pump operation, etc. With model-based methods reaching their limits due to various uncertainty sources, AI methods offer promising solutions to those challenges.
In this work, we introduce a Python toolbox for complex scenario modeling \& generation such that AI researchers can easily access challenging problems from the drinking water domain.
Besides providing a high-level interface for the easy generation of hydraulic and water quality scenario data, it also provides easy access to popular event detection benchmarks and an environment for developing control algorithms.
\end{abstract}

\blfootnote

\section{Introduction}

\begin{figure}
    \centering
    \includegraphics[width=\linewidth]{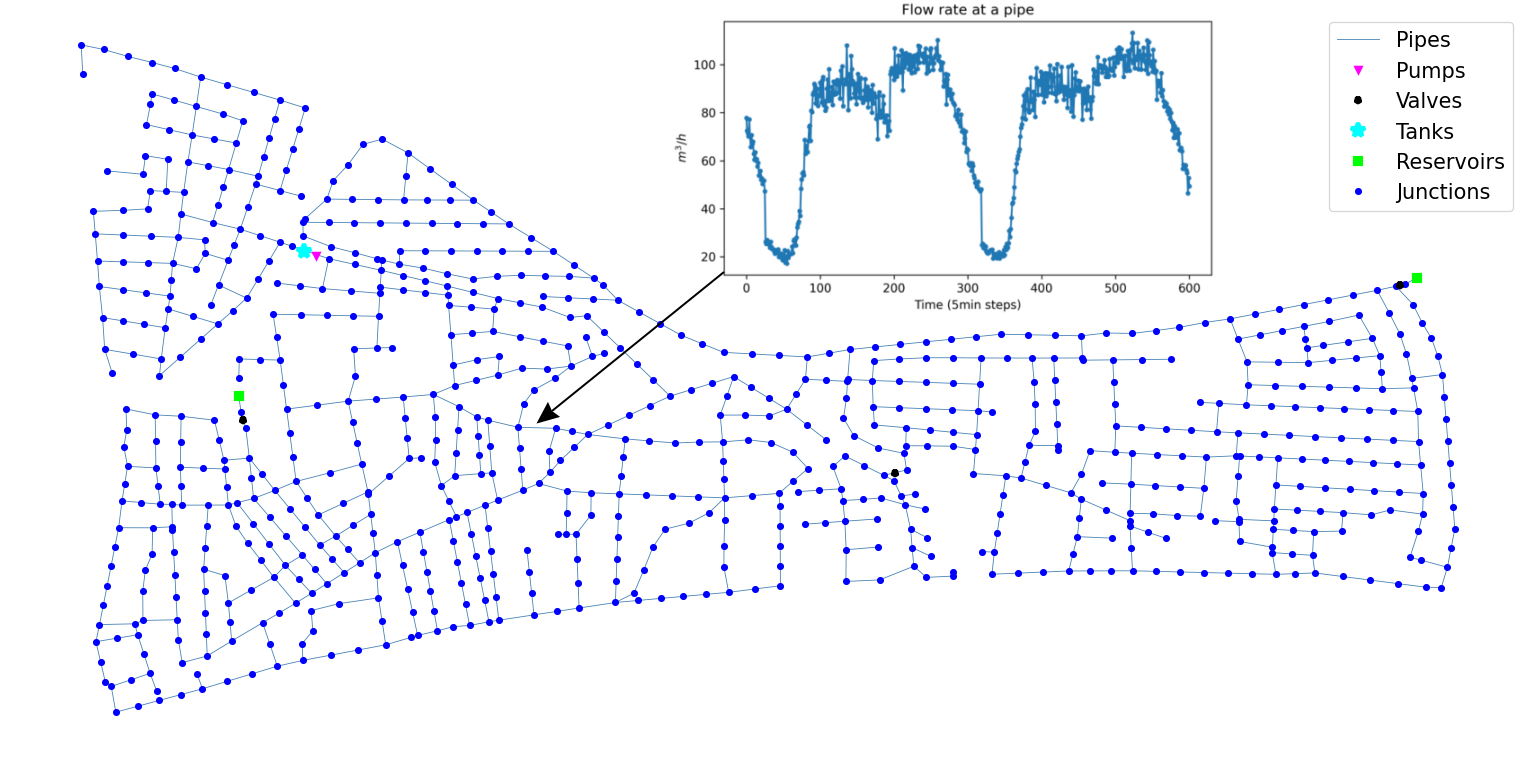}
    \caption{Illustration of the L-Town~\protect\cite{vrachimis2022battledim} Water Distribution Network together with the flow at pipe "p227".}
    \label{fig:wdn_illustration}
\end{figure}

The operation of Water Distribution Networks (WDNs) aims to ensure a reliable supply of drinking water, which is vital for modern societies.
WDNs are typically modeled as graphs where the nodes (i.e. junctions) correspond to consumer demands (e.g. households consuming water) and the edges correspond to pipes. Hydraulic models are used to model water flow and pressure in the network, considering water consumption and operation of pumps and valves. Moreover, water quality is modeled using partial differential equations describing the chemical reaction and transport of different substances through the network.
The schematics of a WDN are illustrated in Figure~\ref{fig:wdn_illustration}.
In practice, WDNs are operated and monitored by humans, which are supported by software, such as basic control algorithms and event detectors, relying on a few sensors in the WDN. However, given the rapid growth of urban areas, WDNs are becoming more complex, and modeling uncertainties prevent their efficient operation.
Besides classic event detection and isolation (e.g. leakage detection), control tasks such as pump scheduling, as well as modeling/prediction/forecasting of water quality states in WDNs are becoming more challenging.
Climate change and rapid population growth significantly affect the quality of source water and the ability of WDN operators to predict and detect water quality issues.
With increasing and time-varying uncertainty, new AI and data-driven methods provide promising tools for tackling such challenges.
Currently, non-water experts such as AI researchers face several challenges when devising practical solutions for water system applications, such as the unavailability of tools for easy scenario/data generation and easy access to benchmarks, that hinder the progress of applying AI to this domain.
Easy-to-use toolboxes and access to benchmark data sets are extremely important for boosting and accelerating research, as well as for supporting reproducible research, as it was, for instance, the case in deep learning and machine learning where toolboxes such as TensorFlow and scikit-learn had a significant impact on boosting research.

The contributions of this paper are two-fold: First, we introduce AI researchers to emerging problems in the drinking water domain where AI can have a significant impact. Second, we provide a toolbox for generating and accessing data to build AI-based solutions for those challenges.

\section{Related Work}

The modeling and simulation of hydraulic and water quality dynamics in water distribution networks have progressively advanced with the introduction of simulation software. 
Notably, EPANET~\cite{rossman2000} and its extension EPANET-MSX~\cite{shang2008modeling} are foundational tools in this area.
These are complemented by tools that make use of high-level programming languages, such as the EPANET-MATLAB Toolkit (EMT) ~\cite{eliades2016}, the Object-Oriented Pipe Network Analyzer (OOPNET) \cite{Steffelbauer2015}, and the EPANET-Python Toolkit (EPyT)~\cite{kyriakou2023epyt}. 
These tools are instrumental in facilitating research into WDN resilience and response to various operational challenges.
%
%
These tools, however, lack support for the creation of realistic (benchmark) scenarios by missing implementations of essential aspects such as realistic fault models (of leakages and sensor faults), various sensor configurations, custom control modules, and other events such as changes in water quality caused by external factors.
A first step towards such software for scenario creation is the  Water Network Tool for Resilience (WNTR)~\cite{klise2017software}, which facilitates the simulation of hydraulic dynamics, and in addition, it allows the simulation of various events such as pipe breaks, disasters such as earthquakes, power outages, and fires. However, it currently does not support quality dynamics and also misses other crucial aspects such as sensor configurations, and industrial control modules.
%
The creation and adoption of standardized benchmarks, like LeakDB~\cite{vrachimis2018leakdb} and BattLeDIM~\cite{vrachimis2022battledim} represent significant strides toward enabling comparative studies and enhancing reproducibility in the field. LeakDB, created using the EMT, offers a rich set of realistic leakage scenarios, aiding in the development and evaluation of anomaly detection algorithms. Similarly, BattLeDIM provides a dataset for benchmarking algorithms' ability to detect and isolate (i.e., localize) leakages, contributing valuable insights into the effectiveness of different methodological approaches.
However, several challenges remain unaddressed. The field lacks easy accessibility to realistic datasets that encapsulate the complexity of real-world water distribution systems, limiting the development and benchmarking of algorithms and methodologies. There's a critical need for comprehensive benchmarks and software for scenario generation that can serve as common grounds for developing, testing, and evaluating algorithms. 
Reflecting on the state of the field, it is evident that, while existing tools have provided significant capabilities, the integration of realistic scenario generation with real-time data acquisition and advanced analytics into these platforms is crucial for the next leap in water network management and research.

\section{AI in Water Distribution Networks}

Artificial Intelligence (AI) has emerged as a key tool for WDNs, providing potential solutions for water demand management, leak detection, pump optimization, and infrastructure resilience.
AI methodologies like machine learning and deep learning excel at analyzing complex data from diverse sources. 
This ability is especially useful in WDNs where uncertainties such as fluctuating demand, variable water quality supply conditions, increasing infrastructure complexity, and a relatively small number of measuring devices, prohibit the extraction of valuable predictive insights and risk assessment.
AI applications extend to the development of surrogate models \cite{Zanfei2023,Ashraf_Strotherm_Hermes_Hammer_2024}, which speed up complex simulations and enable real-time decision support.
Moreover, AI can assist in modeling the complex dynamics of water quality and predicting potential issues, enabling proactive actions~\cite{Li2024}. 
One of the critical applications of AI is in the detection, isolation, and localization of events like leakages.
A recent example \cite{Rajabi2023} investigates the application of conditional convolutional generative adversarial networks for real-time leak detection and localization using hydraulic model-based image conversion and the structural similarity index.
Moreover, events such as contamination \cite{Li2023}, and cyber-physical attacks \cite{Taormina2018a} have also been benefited by AI-driven systems that analyze sensor data in real-time, detecting deviations that may indicate such events, thereby facilitating rapid and informed responses to mitigate impacts on public health and network integrity.
Additionally, AI can enhance the control of network components like pumps and valves, optimizing their operation based on predictive models and real-time data. Such AI-enhanced control systems may be able to dynamically adjust operational parameters, improving resource efficiency, and reducing costs \cite{Guo2023}.

\section{EPyT-Flow: A Python Toolbox}
\begin{figure}
    \centering
    \includegraphics[width=\linewidth]{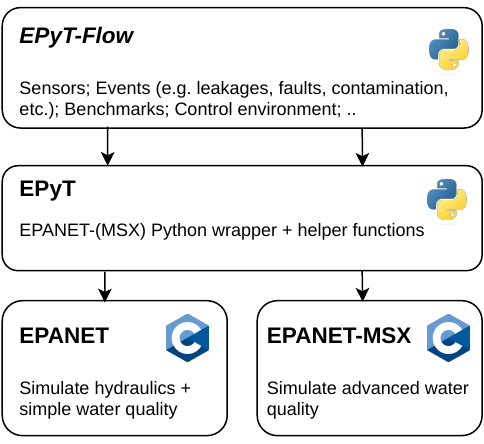}
    \caption{Illustration of the functionality of the proposed toolbox \emph{EPyT-Flow}.}
    \label{fig:toolbox:structure}
\end{figure}

In this paper, we propose a Python toolbox called \emph{EPyT-Flow} that provides a high-level interface for the easy generation of WDN scenario data, but also provides access to low-level functions for maximum flexibility as needed by domain experts -- by this we aim to satisfy different needs and make it a toolbox for "everybody". In addition, its design and object-orientated implementation makes it easy to customize existing functionalities and implement new ones.
Our proposed toolbox \emph{EPyT-Flow} builds upon EPyT which itself provides a Python interface to EPANET and EPANET-MSX -- see Figure~\ref{fig:toolbox:structure} for an illustration.
The toolbox currently includes $16$ WDNs that can be used for scenario generation. However, it also goes beyond pure scenario generation by providing access to $7$ popular and widely adopted benchmarks on event detection and localization (incl. their evaluation metrics) -- ready to be utilized for building and evaluating AI models.
It also comes with an implemented residual-based sensor interpolation event detection method that can be used as a baseline for any event detection benchmark.
Furthermore, it also provides an environment (inspired by the OpenAI Gym) for developing and implementing control algorithms -- i.e. a gym for training reinforcement learning agents to solve some given control tasks such as energy efficient pump scheduling.
In order to support modeling a wide variety of scenarios, the toolbox comes with $4$ different event types and a total number of $13$ pre-defined and implemented events ready to be utilized in custom scenarios: $3$ different leakages, $3$ actuator events, $5$ sensor faults, and $2$ communication events. Furthermore, all those events can be easily customized as needed by the user.
Since the quantities in the real world are always subject to uncertainty, the toolbox comes with $11$ pre-defined types of uncertainties ranging from classic Gaussian noise to different types of (very) deep uncertainties that can be applied to hydraulic parameters such as pipe length, diameter, and roughness, water quality parameters, such as reaction coefficients, and sensor noise.
%
%
The proposed toolbox (incl. the detailed documentation) is available at \url{https://github.com/WaterFutures/EPyT-Flow}.

\section{Use-Case -- Event Detection in WDNs}
Here, we demonstrate how the proposed toolbox can be utilized to easily set up an event detection scenario and apply a baseline event detection method.

We load the popular L-Town network~\cite{vrachimis2022battledim} ($785$ nodes and $909$ pipes -- see Figure~\ref{fig:wdn_illustration}) with the default sensor placement from the BattLeDIM competition ($29$ pressure and $2$ flow sensors) and demand patterns for modeling a realistic water consumption behavior:
\begin{lstlisting}[language=Python]
cfg = load_ltown(use_realistic_demands=True,
        include_default_sensor_placement=True)
\end{lstlisting}
We create a new scenario ($14$ days long with $5$min time steps) based on this configuration:
\begin{lstlisting}[language=Python]
sim = ScenarioSimulator(scenario_config=cfg)
params = {
 "simulation_duration": to_seconds(days=14),
 "hydraulic_time_step": to_seconds(minutes=5)}
sim.set_general_parameters(**params)
\end{lstlisting}
The first week of data in this scenario does not contain any events, in the second week, however, we add a small abrupt leakage, and a slightly larger incipient leakage:
\begin{lstlisting}[language=Python]
sim.add_leakage(AbruptLeakage(link_id="p673",
            diameter=0.001,
            start_time=to_seconds(days=7),
            end_time=to_seconds(days=8)))
sim.add_leakage(IncipientLeakage(link_id="p31",
            diameter=0.02,
            start_time=to_seconds(days=11),
            end_time=to_seconds(days=13),
            peak_time=to_seconds(days=12)))
\end{lstlisting}
Moreover, a flow sensor fault is included:
\begin{lstlisting}[language=Python]
sim.add_sensor_fault(SensorFaultDrift(coef=1.1,
        sensor_id="p227",
        sensor_type=SENSOR_TYPE_LINK_FLOW,
        start_time=to_seconds(days=9),
        end_time=to_seconds(days=10)))
\end{lstlisting}
Finally, we run the simulation to obtain the data (i.e. sensor readings) as they would be received in an operator's Supervisory Control and Data Acquisition (SCADA) system, and are processed by the event detector:
\begin{lstlisting}[language=Python]
scada_data = sim.run_simulation()
\end{lstlisting}
As a classic baseline, our proposed toolbox already implements a residual-based interpolation detection method -- this method tries to predict the readings of a given sensor based on all other sensors: $f: \vec{x}_t\setminus\{i\} \mapsto (\vec{x}_t)_i$, where $\vec{x}_t$ refers to these sensor ratings at time $t$, and $\vec{x}_t\setminus\{i\}$ denotes these sensor readings without the $i$-th sensor.
An alarm is raised (i.e. event detected) whenever the prediction and the observation of at least one sensor differ significantly:
\begin{equation}
   \exists i:\; |f(\vec{x}_t\setminus\{i\}) - (\vec{x}_t)_i| > \theta_i
\end{equation}
where $\theta_i > 0$ denotes a sensor-specific threshold at which the difference is considered as significant.
For this, the detection method has to be calibrated (i.e. fitted) to a time window of (ideally event-free) sensor readings to determine a suitable threshold $\theta$ that does not raise an alarm when the network is in normal operation (i.e. no events present).
We split the simulated data into half -- the first half corresponds to the first week where no events are active and the second half contains some leakages and sensor faults as described above.
\begin{lstlisting}[language=Python]
X = scada_data.get_data()  # 4000 time steps
X_train, X_test = X[:2000, :], X[2000:, :]
\end{lstlisting}
The first half of the data is event-free and is used to calibrate (i.e. fit) the event detector:
\begin{lstlisting}[language=Python]
detector = SensorInterpolationDetector()
detector.fit(X_train)
\end{lstlisting}
, which then is applied to the second half in order to detect suspicious time points where an event might have been active:
\begin{lstlisting}[language=Python]
suspicious_time_points = detector.apply(X_test)
\end{lstlisting}
\begin{figure}
    \centering
    \includegraphics[width=\linewidth]{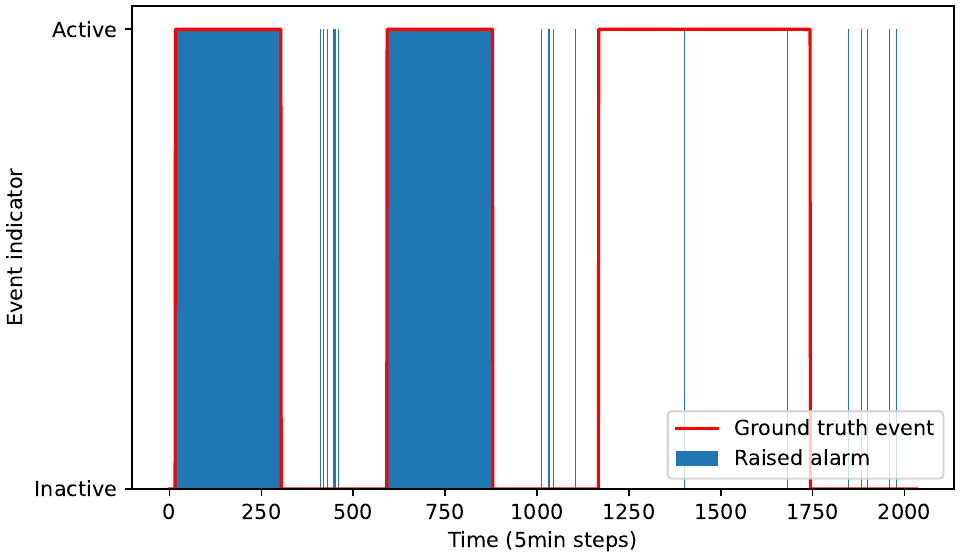}
    \caption{Evaluation of the event detection on the second week of the simulated scenario.}
    \label{fig:results}
\end{figure}
The results are visualized in Figure~\ref{fig:results}. The event detection method does a good job of detecting the abrupt leakage (first event), as well as the sensor fault (second event). However, it fails in case of the incipient leakage -- note that incipient leakages are expected to be more challenging due to the gradual change in the sensor readings instead of an abrupt change. Also, the method yields a few false positives as well.

An AI researcher could develop and apply more sophisticated event detection methods and compare their performance against this baseline.

\section{Conclusion \& Summary}

In this work, we introduced a Python toolbox for realistic scenario data generation and access to benchmarks of WDNs, that AI researchers can utilize to develop methods to support human WDN operators in various real-world challenges.

Our long-term vision for this toolbox is to split it into three parts to further facilitate the unfolding of the huge potential of AI in WDNs:
1) A core part for data generation (i.e., scenario simulation);
2) A \emph{BenchmarkHub} as a platform for accessing and sharing WDN benchmarks;
3) A \emph{ModelHub} as a platform for accessing and sharing AI \& classic models and algorithms for different tasks in WDNs.

\section*{Acknowledgments}
This research was supported by the Ministry of Culture and Science NRW (Germany) as part of the Lamarr Fellow Network.
We also gratefully acknowledge funding from the European Research Council (ERC) under the ERC Synergy Grant Water-Futures (Grant agreement No. 951424).
This publication reflects the views of the authors only.

\bibliographystyle{named}
\bibliography{ijcai24}

\end{document}